\documentclass[11pt,a4paper]{article}
\usepackage[hyperref]{acl2019}
\usepackage{times}
\usepackage{latexsym}

\usepackage[utf8]{inputenc}
\usepackage{amsmath}
\usepackage{amsfonts}
\usepackage{graphicx}
\usepackage{booktabs}
\usepackage{multirow}
\usepackage{enumitem}
\usepackage[colorinlistoftodos]{todonotes}

\usepackage{subfigure}
\usepackage{textcomp}
\usepackage{latexsym}
\usepackage{graphicx}
\usepackage{url}

\aclfinalcopy % Uncomment this line for the final submission
\def\aclpaperid{2196} %  Enter the acl Paper ID here

\title{\textsc{TweetQA}: A Social Media Focused Question Answering Dataset}

\author{
 Wenhan Xiong$^\dagger$,
 Jiawei Wu$^\dagger$,
 Hong Wang$^\dagger$,
 Vivek Kulkarni$^\dagger$,\\
 \bf Mo Yu$^\ast$, 
 Shiyu Chang$^\ast$, 
 Xiaoxiao Guo$^\ast$, 
 William Yang Wang$^\dagger$
\\ 
 $^\dagger$ University of California, Santa Barbara\\
 $^\ast$ IBM Research\\
 \{xwhan, william\}@cs.ucsb.edu, yum@us.ibm.com, \{shiyu.chang, xiaoxiao.guo\}@ibm.com  
 }
 
\date{}
\date{}

\begin{document}
\maketitle
\begin{abstract}
With social media becoming increasingly popular on which lots of news and real-time events are reported, developing automated question answering systems is critical to the effectiveness of many applications that rely on real-time knowledge. While previous datasets have concentrated on question answering (QA) for formal text like news and Wikipedia, we present the first large-scale dataset for QA over social media data. To ensure that the tweets we collected are useful, we only gather tweets used by journalists to write news articles. We then ask human annotators to write questions and answers upon these tweets. Unlike other QA datasets like SQuAD in which the answers are extractive, we allow the answers to be abstractive. We show that two recently proposed neural models that perform well on formal texts are limited in their performance when applied to our dataset. In addition, even the fine-tuned BERT model is still lagging behind human performance with a large margin.  Our results thus point to the need of improved QA systems targeting social media text.~\footnote{The Dataset can be found at \url{https://tweetqa.github.io/}.}
\end{abstract}

\section{Introduction}
Social media is now becoming an important real-time information source, especially during natural disasters and emergencies. It is now very common for traditional news media to frequently probe users and resort to social media platforms to obtain real-time developments of events. According to a recent survey by Pew Research Center\footnote{http://www.journalism.org/2017/09/07/news-use-across-social-media-platforms-2017/}, in 2017, more than two-thirds of Americans read some of their news on social media. Even for American people who are 50 or older, $55\%$ of them report getting news from social media, which is $10\%$ points higher than the number in 2016. Among all major social media sites, Twitter is most frequently used as a news source, with $74\%$ of its users obtaining their news from Twitter.  All these statistical facts suggest that understanding user-generated noisy social media text from Twitter is a significant task. 

\begin{table}[t]
\centering
\begin{tabular}{ll}
\toprule
\multicolumn{2}{p{7cm}}{\textbf{Passage}: \textit{Oh man just read about Paul Walkers death. So young. Ugggh makes me sick especially when it's caused by an accident. God bless his soul. -- \textcolor{red}{Jay Sean (@jaysean)} \textcolor{blue}{December 1, 2013}}} \\
\midrule
\textbf{Q}: \textit{why is sean torn over the actor's death?} \\
\textbf{A}: \textit{walker was young}\\
\bottomrule
\end{tabular}
\caption{An example showing challenges of \textsc{TweetQA}. Note the highly informal nature of the text and the presence of social media specific text like usernames which need to be comprehended to accurately answer the question.}
\label{tab:example}
\end{table}

In recent years, while several tools for core natural language understanding tasks involving syntactic and semantic analysis have been developed for noisy social media text~\cite{Gimpel2011PartofSpeechTF,ritter2011named,kong2014dependency,Wang:2014:EMNLP}, there is little work on question answering or reading comprehension over social media, with the primary bottleneck being the lack of available datasets. We observe that recently proposed QA datasets usually focus on formal domains, \emph{e.g.} \textsc{CNN/DailyMail}~\cite{hermann:nips2015} and NewsQA~\cite{trischler2016newsqa} on news articles; SQuAD~\cite{rajpurkar2016squad} and \textsc{WikiMovies}~\cite{miller2016key} that use Wikipedia.

% Recently proposed datasets for QA such as \textsc{CNN/DailyMail}~\cite{hermann:nips2015} and NewsQA~\cite{trischler2016newsqa} focus on news articles, while SQuAD~\cite{rajpurkar2016squad} and \textsc{WikiMovies}~\cite{miller2016key} include only Wikipedia articles. 

In this paper, we propose the first large-scale dataset for QA over social media data. Rather than naively obtaining tweets from Twitter using the Twitter API\footnote{https://developer.twitter.com/} which can yield irrelevant tweets with no valuable information, we restrict ourselves only to tweets which have been used by journalists in news articles thus implicitly implying that such tweets contain useful and relevant information. To obtain such relevant tweets, we crawled thousands of news articles that include tweet quotations and then employed crowd-sourcing to elicit questions and answers based on these event-aligned tweets. Table \ref{tab:example} gives an example from our \textsc{TweetQA} dataset. It shows that QA over tweets raises challenges not only because of the informal nature of oral-style texts (\emph{e.g.} inferring the answer from multiple short sentences, like the phrase ``so young'' that forms an independent sentence in the example), but also from tweet-specific expressions (such as inferring that it is ``Jay Sean'' feeling sad about Paul's death because he posted the tweet).

Furthermore, we show the distinctive nature of \textsc{TweetQA} by comparing the collected data with traditional QA datasets collected primarily from formal domains.  In particular, we demonstrate empirically that three strong neural models which achieve good performance on formal data do not generalize well to social media data, bringing out challenges to developing QA systems that work well on social media domains.

In summary, our contributions are:
\begin{itemize}
\item We present the first question answering dataset, \textsc{TweetQA}, that focuses on social media context;
\item We conduct extensive analysis of questions and answer tuples derived from social media text and distinguish it from standard question answering datasets constructed from formal-text domains;
\item Finally, we show the challenges of question answering on social media text by quantifying the performance gap between human readers and recently proposed neural models, and also provide insights on the difficulties by analyzing the decomposed performance over different question types.
\end{itemize}

\section{Related Work}
\paragraph{Tweet NLP}
Traditional core NLP research typically focuses on English
newswire datasets such as the Penn Treebank~\cite{marcus1993building}.
In recent years, with the increasing usage of social media platforms, several NLP techniques and datasets for processing social media text have been proposed. For example, \newcite{Gimpel2011PartofSpeechTF} build a Twitter part-of-speech tagger based on 1,827 manually annotated tweets. \newcite{ritter2011named} annotated 800 tweets, and performed an empirical study for part-of-speech tagging and chunking on a new Twitter dataset. They also investigated the task of Twitter Named Entity Recognition, utilizing a dataset of 2,400 annotated tweets. \newcite{kong2014dependency} annotated 929 tweets, and built the first dependency parser for tweets, whereas \newcite{Wang:2014:EMNLP} built the Chinese counterpart based on 1,000 annotated Weibo posts. To the best of our knowledge, question answering and reading comprehension over short and noisy social media data are rarely studied in NLP, and our annotated dataset is also an order of magnitude large than the above public social-media datasets.

%%%%%%%%%%%%%%%%%%%%%%%%%%%%%%%%%%%%%%%%%%%%%%%%%%%%%%%%%%%%%%%%
\paragraph{Reading Comprehension}
Machine reading comprehension (RC) aims to answer questions by comprehending evidence from passages.  This direction has recently drawn much attention due to the fast development of deep learning techniques and large-scale datasets.  The early development of the RC datasets focuses on either the cloze-style \cite{hermann:nips2015,hill2015goldilocks} or quiz-style problems \cite{richardson2013mctest,lai2017race}. 
The former one aims to generate single-token answers from automatically constructed pseudo-questions while the latter requires choosing from multiple answer candidates. However, such unnatural settings make them fail to serve as the standard QA benchmarks. Instead, researchers started to ask human annotators to create questions and answers given passages in a crowdsourced way. Such efforts give the rise of large-scale human-annotated RC datasets, many of which are quite popular in the community such as SQuAD~\cite{rajpurkar2016squad}, MS MARCO~\cite{nguyen2016ms}, NewsQA~\cite{trischler2016newsqa}. More recently, researchers propose even challenging datasets that require QA within dialogue or conversational context~\cite{reddy2018coqa,choi2018quac}. According to the difference of the answer format, these datasets can be further divided to two major categories: \emph{extractive} and \emph{abstractive}.  In the first category, the answers are in text spans of the given passages, while in the latter case, the answers may not appear in the passages.  It is worth mentioning that in almost all previously developed datasets, the passages are from Wikipedia, news articles or fiction stories, which are considered as the formal language.  Yet, there is little effort on RC over informal one like tweets.

%%%%%%%%%%%%%%%%%%%%%%%%%%%%%%%%%%%%%%%%%%%%%%%%%%%%%%%%%%%%%%%%
\section{TweetQA}
In this section, we first describe the three-step data collection process of \textsc{TweetQA}: \textit{tweet crawling}, \textit{question-answer writing} and \textit{answer validation}. Next, we define the specific task of \textsc{TweetQA} and discuss several evaluation metrics. To better understand the characteristics of the \textsc{TweetQA} task, we also include our analysis on the answer and question characteristics using a subset of QA pairs from the development set.

%%%%%%%%%%%%%%%%%%%%%%%%%%%%%%%%%%%%%%%%%%%%%%%%%%%%%%%%%%%%%%%%%%%%%%%%
\subsection{Data Collection}

\paragraph{Tweet Crawling}
One major challenge of building a QA dataset on tweets is the sparsity of informative tweets. Many users write tweets to express their feelings or emotions about their personal lives. These tweets are generally uninformative and also very difficult to ask questions about. Given the linguistic variance of tweets, it is generally hard to directly distinguish those tweets from informative ones. In terms of this, rather than starting from Twitter API Search, we look into the archived snapshots\footnote{https://archive.org/} of two major news websites (CNN, NBC), and then extract the tweet blocks that are embedded in the news articles. In order to get enough data, we first extract the URLs of all section pages (\emph{e.g.} World, Politics, Money, Tech) from the snapshot of each home page and then crawl all articles with tweets from these section pages. Note that another possible way to collect informative tweets is to download the tweets that are posted by the official Twitter accounts of news media. However, these tweets are often just the summaries of news articles, which are written in formal text. As our focus is to develop a dataset for QA on informal social media text, we do not consider this approach. 

After we extracted tweets from archived news articles, we observed that there is still a portion of tweets that have very simple semantic structures and thus are very difficult to raise meaningful questions. An example of such tweets can be like: \textit{``Wanted to share this today - @IAmSteveHarvey"}. This tweet is actually talking about an image attached to this tweet. Some other tweets with simple text structures may talk about an inserted link or even videos. To filter out these tweets that heavily rely on attached media to convey information, we utilize a state-of-the-art semantic role labeling model trained on CoNLL-2005~\cite{He2017DeepSR} to analyze the predicate-argument structure of the tweets collected from news articles and keep only the tweets with more than two labeled arguments. This filtering process also automatically filters out most of the short tweets. For the tweets collected from CNN, $22.8\%$ of them were filtered via semantic role labeling. For tweets from NBC, $24.1\%$ of the tweets were filtered.

\paragraph{Question-Answer Writing}

\begin{figure}[t]
\centering
\includegraphics[width=1.0\linewidth]{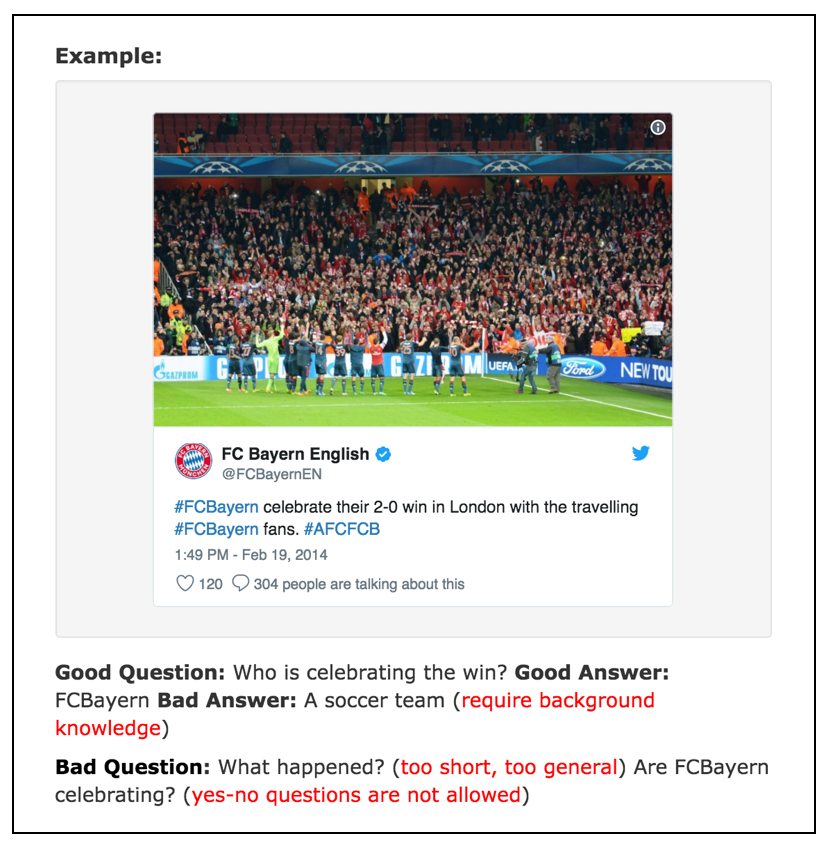}
\caption{An example we use to guide the crowdworkers when eliciting question answer pairs. We elicit question that are neither too specific nor too general, do not require background knowledge.}
\label{ui}
\end{figure}

We then use Amazon Mechanical Turk to collect question-answer pairs for the filtered tweets. For each Human Intelligence Task (HIT), we ask the worker to read three tweets and write two question-answer pairs for each tweet. To ensure the quality, we require the workers to be located in major English speaking countries (\emph{i.e.} Canada, US, and UK) and have an acceptance rate larger than $95\%$. Since we use tweets as context, lots of important information are contained in hashtags or even emojis. Instead of only showing the text to the workers, we use \textit{javascript} to directly embed the whole tweet into each HIT. This gives workers the same experience as reading tweets via web browsers and help them to better compose questions.

To avoid trivial questions that can be simply answered by superficial text matching methods or too challenging questions that require background knowledge. We explicitly state the following items in the HIT instructions for question writing:
\begin{itemize}
\item No Yes-no questions should be asked.
\item The question should have at least five words.
\item Videos, images or inserted links should not be considered.
\item No background knowledge should be required to answer the question.
\end{itemize}
To help the workers better follow the instructions, we also include a representative example showing both good and bad questions or answers in our instructions. Figure \ref{ui} shows the example we use to guide the workers. 

As for the answers, since the context we consider is relatively shorter than the context of previous datasets, we do not restrict the answers to be in the tweet, otherwise, the task may potentially be simplified as a classification problem. The workers are allowed to write their answers in their own words. We just require the answers to be brief and can be directly inferred from the tweets. 

After we retrieve the QA pairs from all HITs, we conduct further post-filtering to filter out the pairs from workers that obviously do not follow instructions. We remove QA pairs with yes/no answers. Questions with less than five words are also filtered out. This process filtered $13\%$ of the QA pairs. The dataset now includes 10,898 articles, 17,794 tweets, and 13,757 crowdsourced question-answer pairs. The collected QA pairs will be directly available to the public,
and we will provide a script to download the original tweets and detailed documentation on how we build our dataset. Also note that since we keep the original news article and news titles for each tweet, our dataset can also be used to explore more challenging generation tasks.
 Table~\ref{stats} shows the statistics of our current collection, and the frequency of different types of questions is shown in Table~\ref{q_types}. All QA pairs were written by 492 individual workers. 

\begin{table}[t]
\centering
\small
\begin{tabular}{l|c}
\toprule
\multicolumn{2}{c}{\textbf{Dataset Statistics}}\\
\midrule
\# of Training triples & 10,692  \\
\# of Development triples & 1,086 \\
\# of Test triples & 1,979\\
\midrule
Average question length (\#words) & 6.95\\
Average answer length (\#words) & 2.45\\
\bottomrule
\end{tabular}
\caption{Basic statistics of \textsc{TweetQA}}
\label{stats}
\end{table}

\paragraph{Answer Validation}

\begin{table}[t]
\small
\centering
\begin{tabular}{l|c}
\toprule
\textbf{Question Type} & \textbf{Percentage}\\
\midrule
What & 42.33\%  \\
Who & 29.36\% \\
How & 7.79\% \\
Where & 7.00\% \\
Why & 2.61\% \\
Which & 2.43\% \\
When & 2.16\% \\
Others & 6.32\% \\
\bottomrule
\end{tabular}
\caption{Question Type statistics of \textsc{TweetQA}}
\label{q_types}
\end{table}

For the purposes of human performance evaluation and inter-annotator agreement checking, we launch a different set of HITs to ask workers to answer questions in the test and development set. The workers are shown with the tweet blocks as well as the questions collected in the previous step. At this step, workers are allowed to label the questions as ``NA" if they think the questions are not answerable. We find that $3.1\%$ of the questions are labeled as unanswerable by the workers (for SQuAD, the ratio is $2.6\%$). Since the answers collected at this step and previous step are written by different workers, the answers can be written in different text forms even they are semantically equal to each other. For example, one answer can be ``\textit{Hillary Clinton}'' while the other is ``\textit{@HillaryClinton}''. As it is not straightforward to automatically calculate the overall agreement, we manually check the agreement on a subset of 200 random samples from the development set and ask an independent human moderator to verify the result. It turns out that $90\%$ of the answers pairs are semantically equivalent, $2\%$ of them are partially equivalent (one of them is incomplete) and $8\%$ are totally inconsistent. The answers collected at this step are also used to measure the human performance. We have 59 individual workers participated in this process.

%%%%%%%%%%%%%%%%%%%%%%%%%%%%%%%%%%%%%%%%%%%%%%%%%%%%%%%%%%%%%%%%%%%%%%%%
\subsection{Task and Evaluation}
\label{sec:eval}
As described in the \textit{question-answer writing} process, the answers in our dataset are different from those in some existing extractive datasets. Thus we consider the task of answer generation for \textsc{TweetQA} and we use several standard metrics for natural language generation to evaluate QA systems on our dataset, namely we consider BLEU-1\footnote{The answer phrases in our dataset are relatively short so we do not consider other BLEU scores in our experiments}~\cite{Papineni2002BleuAM}, Meteor~\cite{Denkowski2011Meteor1A} and Rouge-L~\cite{Lin2004ROUGEAP} in this paper. 
% As the answers to questions on tweets can be relatively shorter than those in the previous dataset, character-level metrics (treat each character as unigram) may also be considered. We include the scores of these character-level metrics in the appendix.

To evaluate machine systems, we compute the scores using both the original answer and validation answer as references. For human performance, we use the validation answers as generated ones and the original answers as references to calculate the scores.

%%%%%%%%%%%%%%%%%%%%%%%%%%%%%%%%%%%%%%%%%%%%%%%%%%%%%%%%%%%%%%%%%%%%%%%%
\subsection{Analysis}
% \paragraph{Types of Reasoning required to Answer Questions}

\begin{figure*}
\centering
\includegraphics[width=1.0\linewidth]{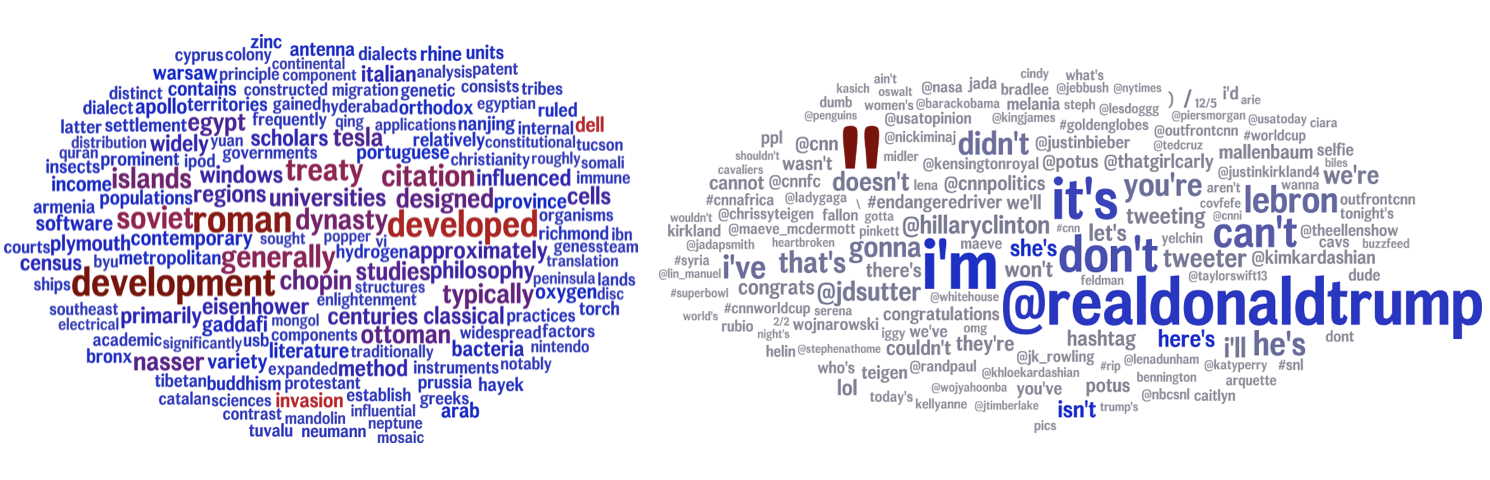}
\caption{Visualization of vocabulary differences between SQuAD (left) and \textsc{TweetQA} (right). Note the presence of a heavy tail of hash-tags and usernames on \textsc{TweetQA} that are rarely found on SQuAD. The color range from red to gray indicates the frequency (red the highest and gray the lowest).}
\label{embed}
\end{figure*}

\begin{table*}[thbp]
\centering
\begin{small}
\begin{tabular}{l|c|l}
\toprule
\bf Type & \bf Fraction (\%) & \bf Example \\
\midrule
\multirow{5}{*}{Paraphrasing only} &  \multirow{5}{*}{47.3} & \multirow{3}{9cm}{P: \emph{Belgium camp is 32 miles from canceled game at US base. Surprised Klinsmann didn't offer to use his helicopter pilot skills to give a ride. -- Grant Wahl (@GrantWahl)}} \\
& & \\
& & \\
& & Q: \emph{what expertise does klinsmann possess?} \\
&& A: \emph{helicopter pilot skills} \\
\midrule
\midrule
\multicolumn{3}{c}{\underline{Types Beyond Paraphrasing}} \\
\multirow{5}{*}{Sentence relations} & \multirow{5}{*}{10.7}& \multirow{3}{9cm}{P: \emph{My heart is hurting. You were an amazing tv daddy! Proud and honored to have worked with one of the best. Love and Prayers \#DavidCassidy— Alexa PenaVega (@alexavega) November 22, 2017}} \\
& & \\
& & \\
&& Q: \emph{who was an amazing tv daddy?}\\
&& A: \emph{\#davidcassidy}\\
\midrule
\multirow{5}{*}{\underline{Authorship}}& \multirow{5}{*}{17.3} & \multirow{3}{9cm}{P: \emph{Oh man just read about Paul Walkers death. So young. Ugggh makes me sick  especially when it's caused by an accident. God bless his soul. -- Jay Sean (@jaysean)}} \\
&&\\
& & \\
& & Q: \emph{why is sean torn over the actor's death?} \\
&& A: \emph{walker was young} \\
\midrule
\multirow{4}{*}{\underline{Oral/Tweet English habits}} & \multirow{5}{*}{10.7} & \multirow{2}{9cm}{P: \emph{I got two ways to watch the OLYMPICS!! CHEAH!! USA!! Leslie Jones (@Lesdoggg) August 6, 2016}} \\
&&\\
& & Q: \emph{who is being cheered for?}\\
& & A: \emph{usa}\\
\midrule
\multirow{5}{*}{\underline{UserIDs \& Hashtags}} & \multirow{4}{*}{12.0} & \multirow{3}{9cm}{P: \emph{Started researching this novel in 2009. Now it is almost ready for you to read. Excited! \#InTheUnlikelyEvent -- Judy Blume (@judyblume)}}\\
&& \\
&& \\
& & Q: \emph{what is the name of the novel?} \\
& & A: \emph{in the unlikely event.} \\
\midrule
\multirow{4}{*}{Other commonsense} & \multirow{4}{*}{6.7} & \multirow{2}{9cm}{P: \emph{Don't have to be Sherlock Holmes to figure out what Russia is up to ... -- Lindsey Graham (@LindseyGrahamSC)}}\\
&&\\
&& Q: \emph{what literary character is referenced?} \\
&& A: \emph{sherlock holmes.}\\
\midrule
\multirow{4}{*}{Deep semantic} & \multirow{4}{*}{3.3} & \multirow{2}{9cm}{P: \emph{@MayorMark its all fun and games now wait until we are old enough to vote \#lastlaugh -- Dylan (@DFPFilms1)}}\\
&&\\
&& Q: \emph{when does the author suggest a change?}\\
&& A: \emph{when he's of voting age.}\\
\midrule
\multirow{5}{*}{Ambiguous} & \multirow{5}{*}{5.3} & \multirow{3}{9cm}{P: \emph{The \#endangeredriver would be a sexy bastard in this channel if it  had water. Quick turns. Narrow. (I'm losing it) -- John D. Sutter (@jdsutter)}}\\
&&\\
&&\\
(Meaningless questions) && Q: \emph{what is this user "losing"}\\
 && A: \emph{he is losing it}\\
\bottomrule
\end{tabular}
\end{small}
\caption{Types of reasoning abilities required by \textsc{TweetQA}. 
\underline{Underline} indicates tweet-specific reasoning types, which are common in \textsc{TweetQA} but are rarely observed in previous QA datasets.
Note that the first type represents questions that \emph{only} require the ability of paraphrasing, while the rest of the types require some other more salient abilities besides paraphrasing. \textbf{Overlaps could exist between different reasoning types in the table}. For example, the \emph{second} example requires both the understanding of sentences relations and tweet language habits to answer the question; and the \emph{third} example requires both the understanding of sentences relations and authorship.}
\label{tab:reasoning_type}
\end{table*}

In this section, we analyze our dataset and outline the key properties that distinguish it from standard QA datasets like SQuAD \cite{rajpurkar2016squad}. First, our dataset is derived from social media text which can be quite informal and user-centric as opposed to SQuAD which is derived from Wikipedia and hence more formal in nature. We observe that the shared vocabulary between SQuAD and \textsc{TweetQA} is only $10.79\%$, suggesting a significant difference in their lexical content. Figure \ref{embed} shows the $1000$ most distinctive words in each domain as extracted from SQuAD and \textsc{TweetQA}.  Note the stark differences in the words seen in the \textsc{TweetQA} dataset, which include a large number of user accounts with a heavy tail. Examples include \texttt{@realdonaldtrump, @jdsutter, @justinkirkland} and \texttt{\#cnnworldcup, \#goldenglobes}. In contrast, the SQuAD dataset rarely has usernames or hashtags that are used to signify events or refer to the authors. It is also worth noting that the data collected from social media can not only capture events and developments in real-time but also capture individual opinions and thus requires reasoning related to the authorship of the content as is illustrated in Table \ref{tab:example}.
In addition, while SQuAD requires all answers to be spans from the given passage, we do not enforce any such restriction and answers can be free-form text. In fact, we observed that $~43\%$ of our QA pairs consists of answers which do not have an exact substring matching with their corresponding passages. All of the above distinguishing factors have implications to existing models which we analyze in upcoming sections.

We conduct analysis on a subset of \textsc{TweetQA} to get a better understanding of the kind of reasoning skills that are required to answer these questions.  We sample 150 questions from the development set, then manually label their reasoning categories. Table \ref{tab:reasoning_type} shows the analysis results. We use some of the categories in SQuAD~\cite{rajpurkar2016squad} and also proposes some tweet-specific reasoning types.

Our first observation is that almost half of the questions only require the ability to identify paraphrases. 
Although most of the ``paraphrasing only'' questions are considered as fairly easy questions, we find that a significant amount (about 3/4) of these questions are asked about event-related topics, such as information about ``who did what to whom, when and where''.
% our \textsc{TweetQA} deals with answer generation from passages, making some of these questions still challenging.
This is actually consistent with our motivation to create \textsc{TweetQA}, as we expect this dataset could be used to develop systems that automatically collect information about real-time events.

Apart from these questions, there are also a group of questions that require understanding common sense, deep semantics (\emph{i.e.} the answers cannot be derived from the literal meanings of the tweets), and relations of sentences\footnote{There are more instances of this reasoning type compared to formal datasets since tweets are usually short sentences.} (including co-reference resolution), which are also appeared in other RC datasets \cite{rajpurkar2016squad}.  On the other hand, the \textsc{TweetQA} also has its unique properties.  Specifically, a significant amount of questions require certain reasoning skills that are specific to social media data:

\begin{itemize}[noitemsep]
\item \textbf{Understanding authorship: } Since tweets are highly personal,  it is critical to understand how questions/tweets related to the authors. 

\item \textbf{Oral English \& Tweet English: } Tweets are often oral and informal. QA over tweets requires the understanding of common oral English. Our \textsc{TweetQA} also requires understanding some tweet-specific English, like conversation-style English.

\item \textbf{Understanding of user IDs \& hashtags:} Tweets often contains user IDs and hashtags, which are single special tokens.  Understanding these special tokens is important to answer person- or event-related questions.  

\end{itemize}

%%%%%%%%%%%%%%%%%%%%%%%%%%%%%%%%%%%%%%%%%%%%%%%%%%%%%%%%%%%%%%%%%%%%%%%%%
\section{Experiments}
To show the challenge of \textrm{TweetQA} for existing approaches, we consider four representative methods as baselines. For data processing, we first remove the URLs in the tweets and then tokenize the QA pairs and tweets using NLTK.\footnote{http://www.nltk.org} This process is consistent for all baselines.

\subsection{Query Matching Baseline}
We first consider a simple query matching baseline similar to the IR baseline in~\newcite{Kocisk2017TheNR}. But instead of only considering several genres of spans as potential answers, we try to match the question with all possible spans in the tweet context and choose the span with the highest BLEU-1 score as the final answer, which follows the method and implementation\footnote{https://github.com/shuohangwang/mprc} of answer span selection for open-domain QA~\cite{wang2017r}. We include this baseline to show that \textsc{TweetQA} is a nontrivial task which cannot be easily solved with superficial text matching.

\subsection{Neural Baselines}
We then explore three typical neural models that perform well on existing formal-text datasets. One takes a generative perspective and learns to decode the answer conditioned on the question and context, while the others learns to extract a text span from the context that best answers the question.
\paragraph{Generative QA}

RNN-based encoder-decoder models~\cite{Cho2014LearningPR,Bahdanau2014NeuralMT} have been widely used for natural language generation tasks. Here we consider
a recently proposed generative model~\cite{Song2017AUQ} that first encodes the context and question into a multi-perspective memory via four different neural matching layers, then decodes the answer using an attention-based model equipped with both copy and coverage mechanisms. The model is trained on our dataset for 15 epochs and we choose the model parameters that achieve the best BLEU-1 score on the development set.

\paragraph{BiDAF}
Unlike the aforementioned generative model, the Bi-Directional Attention Flow (BiDAF)~\cite{Seo2016BidirectionalAF} network learns to directly predict the answer span in the context. BiDAF first utilizes multi-level embedding layers to encode both the question and context, then uses bi-directional attention flow to get a query-aware context representation, which is further modeled by an RNN layer to make the span predictions. Since our \textsc{TweetQA} does not have labeled answer spans as in SQuAD, we need to use the human-written answers to retrieve the answer-span labels for training. To get the approximate answer spans, we consider the same matching approach as in the query matching baseline. But instead of using questions to do matching, we use the human-written answers to get the spans that achieve the best BLEU-1 scores.
 
\paragraph{Fine-Tuning BERT}
This is another extractive RC model that benefits from the recent advance in pretrained general language encoders \cite{peters2018deep,devlin2018bert}. In our work, we select the BERT model \cite{devlin2018bert} which has achieved the best  performance on SQuAD. In our experiments, we use the PyTorch reimplementation\footnote{\url{https://github.com/huggingface/pytorch-pretrained-BERT}} of the uncased base model. The batch size is set as 12 and we fine-tune the model for 2 epochs with learning rate 3e-5.
 
\section{Evaluation}
\label{sec:results}

\subsection{Overall Performance}
We test the performance of all baseline systems using the three generative metrics mentioned in Section \ref{sec:eval}. As shown in Table~\ref{results}, there is a large performance gap between human performance and all baseline methods, including BERT, which has achieved superhuman performance on SQuAD.
This confirms than \textsc{TweetQA} is more challenging than formal-test RC tasks.

We also show the upper bound of the extractive models (denoted as \textsc{Extract-Upper}). In the upper bound method, the answers are defined as n-grams from the tweets that maximize the BLEU-1/METEOR/ROUGE-L compared to the annotated groundtruth.
From the results, we can see that the BERT model still lags behind the upper bound significantly, showing great potential for future research.
It is also interesting to see that the \textsc{Human} performance is slightly worse compared to the upper bound. This indicates (1) the difficulty of our problem also exists for human-beings and (2) for the answer verification process, the workers tend to also extract texts from tweets as answers.

According to the comparison between the two non-pretraining baselines, our generative baseline yields better results than BiDAF. We believe this is largely due to the abstractive nature of our dataset, since the workers can sometimes write the answers using their own words.

% The generative baseline yields better results than BiDAF, despite the fact that BiDAF performs better than the generative model on SQuAD. We believe this is largely due to the abstractive nature of our dataset since the workers can sometimes write the answers using their own words. Also, both neural models perform much better than the simple query matching baseline, suggesting that \textsc{TweetQA} is a nontrivial task. 

\begin{table}[t]
\centering
\small
\begin{tabular}{l|c|c|c}
\toprule
& \multicolumn{3}{c}{\textbf{Evaluation on Dev/Test Data}}\\
\cmidrule{2-4}
\textbf{Models} & \textbf{BLEU-1} & \textbf{METEOR} & \textbf{ROUGE-L} \\ \midrule
\textsc{Human} & $76.4|78.2$ & $63.7|66.7$ & $70.9|73.5$ \\ 
\textsc{Extract-UB} & $79.5|80.3$ & $68.8|69.8$ & $74.3|75.6$ 
\\\midrule
Query-Matching & $30.3|29.4$  & $12.0|12.1$ & $17.0|17.4$ \\ \midrule
\multicolumn{4}{c}{\underline{\emph{Neural Baselines}}}\\
BiDAF & $48.3|48.7$ & $31.6|31.4$ & $38.9|38.6$ \\
Generative & $53.4|53.7$ & $32.1|31.8$ & $39.5|39.0$ \\
BERT & $67.3|69.6$ & $56.9|58.6$ & $62.6|64.1$ \\
\bottomrule
\end{tabular}
\caption{Overall performance of baseline models. \textsc{Extract-UB} refers to our estimation of the upper bound of extractive methods.}
\label{results}
\end{table}

% \subsection{Performance Analysis}
\subsection{Performance Analysis over Human-Labeled Question Types}

% \begin{table}[th]
% \centering
% \small
% \begin{tabular}{l|c|c|c}
% \toprule
% \textbf{Reasoning Types} & \multicolumn{3}{c}{\textbf{Generative$|$BERT}} \\
% \cmidrule{2-4}
% & \textbf{BLEU-1} & \textbf{METEOR} &  \textbf{ROUGE-L}\\
% % \cmidrule{2-4}
% % \textbf{Reasoning Types} & \multicolumn{3}{c}{\textbf{Generative$|$BERT}} \\
% \cmidrule{1-4}
% Paraphrasing & 56.8$|$81.7 & 37.6$|$73.4 & 44.1$|$81.8\\ 
% Sentence relations & 53.4$|$50.0 & 34.0$|$46.1 & 42.2$|$51.1\\ 
% Authorship & 65.4$|$63.0 & 38.4$|$55.9 & 46.1$|$61.9\\
% Oral/Tweet habits & 60.8$|$60.4 & 37.2$|$50.3 & 40.7$|51.0^\dagger$\\
% UserIDs\&Hashtags & $41.5^\diamond|29.3^\dagger$ & $3.8^\diamond|13.0^\dagger$ & $9.9^\diamond | 16.2^\dagger$\\
% Commonsense & $38.1^\diamond|72.9$ & $20.1|63.5$ & $33.1|67.1$\\
% Deep semantics & $53.8|25.0^\dagger$ & $7.19^\diamond|7.1^\dagger$ & $13.4^\diamond|10.3^\dagger$\\
% Ambiguous & $18.1^\diamond|31.6^\dagger$ & $4.1^\diamond|25.0^\dagger$ & $11.0^\diamond|67.1$\\
% \bottomrule
% \end{tabular}
% \caption{BiDAF's and the Generative model's performance on questions that require different types of reasoning. \underline{$^\diamond$} and \underline{$^\dagger$} denote the most difficult three types for Generative and BERT models.}
% \label{tab:reason_results}
% \end{table}

\begin{table}[t]
\centering
\small
\begin{tabular}{l|c|c}
\toprule
\textbf{Reasoning Types} & \multicolumn{2}{c}{\textbf{Generative$|$BERT}} \\
\cmidrule{2-3}
& \textbf{METEOR} &  \textbf{ROUGE-L}\\
% \cmidrule{2-4}
% \textbf{Reasoning Types} & \multicolumn{3}{c}{\textbf{Generative$|$BERT}} \\
\cmidrule{1-3}
Paraphrasing & 37.6$|$73.4 & 44.1$|$81.8\\ 
Sentence relations & 34.0$|$46.1 & 42.2$|$51.1\\ 
Authorship & 38.4$|$55.9 & 46.1$|$61.9\\
Oral/Tweet habits &  37.2$|$50.3 & 40.7$|51.0^\dagger$\\
UserIDs\&Hashtags & $3.8^\diamond|13.0^\dagger$ & $9.9^\diamond | 16.2^\dagger$\\
Commonsense  & $20.1|63.5$ & $33.1|67.1$\\
Deep semantics &  $7.19^\diamond|7.1^\dagger$ & $13.4^\diamond|10.3^\dagger$\\
Ambiguous & $4.1^\diamond|25.0^\dagger$ & $11.0^\diamond|67.1$\\
\bottomrule
\end{tabular}
\caption{BiDAF's and the Generative model's performance on questions that require different types of reasoning. \underline{$^\diamond$} and \underline{$^\dagger$} denote the three most difficult reasoning types for the Generative and the BERT models.}
\label{tab:reason_results}
\end{table}

To better understand the difficulty of the \textsc{TweetQA} task for current neural models, we analyze the decomposed model performance on the different kinds of questions that require different types of reasoning (we tested on the subset which has been used for the analysis in Table \ref{tab:reasoning_type}).
% \paragraph{Performance Analysis over Human-Labeled Question Types}
% In order to look into how the RC models' performance differ when dealing with questions that require different types of reasoning, 
% We report the decomposed performance on the sampled subset that has been used in Table \ref{tab:reasoning_type}.
Table \ref{tab:reason_results} shows the results of the best performed non-pretraining  and pretraining approach, \emph{i.e.}, the generative QA baseline and the fine-tuned BERT. Our full comparison including the BiDAF performance and evaluation on more metrics can be found in Appendix \ref{sec:appA}.
Following previous RC research, we also include analysis on automatically-labeled question types in Appendix \ref{sec:appB}.

As indicated by the results on METEOR and ROUGE-L (also indicated by a third metric, BLEU-1, as shown in Appendix \ref{sec:appA}), both baselines perform worse on questions that require the understanding deep semantics and userID\&hashtags.
The former kind of questions also appear in other benchmarks and is known to be challenging for many current models. The second kind of questions is tweet-specific and is related to specific properties of social media data.
Since both models are designed for formal-text passages and there is no special treatment for understanding user IDs and hashtags, the performance is severely limited on the questions requiring such reasoning abilities.
We believe that good segmentation, disambiguation and linking tools developed by the social media community for processing the userIDs and hashtags will significantly help these question types.

% As indicated by the three metrics, both baselines perform worse on questions that require the understanding of commonsense, deep semantics and userID\&hashtags. The first two are known to be challenging for current models and also appear in other datasets while the third one is tweet-specific. Since these models are designed for formal-text passages and it has no special treatment for understanding user IDs and hashtags, the performance is severely limited on the questions requiring such reasoning abilities. For ambiguous questions, we notice that the generative baseline gets much worse results than BiDAF. We conjecture that although these questions are meaningless, they still have many words that overlapped with the contexts. This can give BiDAF potential advantage over the generative baseline.

\paragraph{On non-pretraining model}
Besides the easy questions requiring mainly paraphrasing skill, we also find that the questions requiring the understanding of authorship and oral/tweet English habits are not very difficult. We think this is due to the reason that, except for these tweet-specific tokens, the rest parts of the questions are rather simple, which may require only simple reasoning skill (\emph{e.g.} paraphrasing).

\paragraph{On pretraining model}
Although BERT was demonstrated to be a powerful tool for reading comprehension, this is the first time a detailed analysis has been done on its reasoning skills.
From the results, the huge improvement of BERT mainly comes from two types. The first is paraphrasing, which is not surprising because that a well pretrained language model is expected to be able to better encode sentences. Thus the derived embedding space could work better for sentence comparison.
The second type is commonsense, which is consistent with the good performance of BERT \cite{devlin2018bert} on SWAG \cite{zellers2018swag}.
We believe that this provides further evidence about the connection between large-scaled deep neural language model and certain kinds of commonsense.

\section{Conclusion}
We present the first dataset for QA on social media data by leveraging news media and crowdsourcing. The proposed dataset informs us of the distinctiveness of social media from formal domains in the context of QA. Specifically, we find that QA on social media requires systems to comprehend social media specific linguistic patterns like informality, hashtags, usernames, and authorship. These distinguishing linguistic factors bring up important problems for the research of QA that currently focuses on formal text. We see our dataset as a first step towards enabling not only a deeper understanding of natural language in social media but also rich applications that can extract essential real-time knowledge from social media.

% Also, by constructing the answers in a generative way, we believe that it is important for the research community to move beyond span selection settings, and consider more open-domain, non-factoid questions in the future. 

% We also envision that this line of work can lead to better natural language generation algorithms for social media text, as well as novel applications such as automated creative writing agents that learn to compose news stories from social media text.

% \appendix
% % \input{appendix}
% \begin{table}[h]
% \centering
% % \small
% \begin{tabular}{lc|c|c}
% \toprule
% & \multicolumn{3}{c}{Evaluation on Dev/Test Data}\\
% \cmidrule{2-4}
% Models & BLEU-2 & BLEU-3 & BLEU-4 \\ \midrule
% \textsc{Human} &  &  & \\ \midrule
% BiDAF & & &  \\
% MPQG & & & \\
% \bottomrule
% \end{tabular}
% \caption{Other BLEU scores}
% \label{append1}
% \end{table}

% \begin{table}[h]
% \centering
% % \small
% \begin{tabular}{lc|c|c}
% \toprule
% & \multicolumn{3}{c}{Evaluation on Dev/Test Data}\\
% \cmidrule{2-4}
% Models & BLEU & METEOR & ROUGE-L \\ \midrule
% \textsc{Human} &  &  & \\ \midrule
% BiDAF & & &  \\
% MPQG & & & \\
% \bottomrule
% \end{tabular}
% \caption{Character-level scores}
% \label{append2}
% \end{table}

% \newpage
\bibliography{emnlp2018}
\bibliographystyle{acl_natbib_nourl}

\clearpage

\appendix

% \clearpage

\begin{table*}[h]
\centering
\small
\begin{tabular}{l|c|c|c}
\toprule
& \textbf{BLEU-1} & \textbf{METEOR} &  \textbf{ROUGE-L}\\
\cmidrule{2-4}
\textbf{Reasoning Types} & \multicolumn{3}{c}{\textbf{BiDAF$|$Generative$|$BERT}} \\
\cmidrule{1-4}
Paraphrasing & 49.1$|$56.8$|$81.7 & 35.4$|$37.6$|$73.4 & 44.5$|$44.1$|$81.8\\ 
Sentence relations & 43.3$|$53.4$|$50.0 & 26.8$|$34.0$|$46.1 & 32.8$|$42.2$|$51.1\\ 
Authorship & 52.5$|$65.4$|$63.0 & 30.5$|$38.4$|$55.9 & 42.3$|$46.1$|$61.9\\
Oral/Tweet habits & 45.8$|$60.8$|$60.4 & 34.8$|$37.2$|$50.3 & 35.1$|$40.7$|$51.0$^\dagger$\\
UserIDs\&Hashtags & $30.0^\star | 41.5^\diamond|29.3^\dagger$ & $8.30^\star | 3.81^\diamond|13.0^\dagger$ & $13.7^\star | 9.88^\diamond | 16.2^\dagger$\\
Commonsense & $27.6^\star|38.1^\diamond|72.9$ & $22.4^\star|20.1|63.5$ & $31.0^\star|33.1|67.1$\\
Deep semantics & $34.8^\star|53.8|25.0^\dagger$ & $7.85^\star|7.19^\diamond|7.1^\dagger$ & $17.5^\star|13.4^\diamond|10.3^\dagger$\\
Ambiguous & $35.1|18.1^\diamond|31.6^\dagger$ & $29.2|4.11^\diamond|25.0^\dagger$ & $34.3|11.0^\diamond|67.1$\\
\bottomrule
\end{tabular}
\caption{BiDAF's and the Generative model's performance on questions that require different types of reasoning. \underline{$^\star$}, \underline{$^\diamond$} and \underline{$^\dagger$} denote the three most difficult reasoning types for BiDAF/Generative/BERT models.}
\label{tab:reason_results_full}
\end{table*}

\begin{table*}[th]
\centering
\small
\begin{tabular}{lc|c|c|c|c|c|c|c}
\toprule
& \multicolumn{8}{c}{\textbf{First-Word Question Types}}\\
\cmidrule{2-9}
\textbf{Models} & What & Who & How & Where & When & Why & Which & Others \\ \midrule
\textsc{Human} & 74.1 & 83.5 & 61.1 & 74.8 & 72.2 & 66.0 & 76.8 & 76.0 \\ \midrule
Query-Matching & 32.4 & 29.8 & 28.4 & 27.1 & 22.9 & 51.9 & 22.7 & 21.1 \\ \midrule
\multicolumn{9}{c}{\underline{\emph{Neural Baselines}}}\\
BiDAF & 44.5 & 54.9 & 41.0 & 60.2 & 46.5 & 36.1 & 44.7 & 41.6\\
Generative & 46.8 & 63.8 & 53.4 & 61.7 & 45.4 & 44.3 & 51.4 & 43.1\\
BERT & 64.8 & 72.5 & 57.7 & 78.1 & 64.5 & 61.0 & 67.2 & 59.2 \\
\bottomrule
\end{tabular}
\caption{BLEU-1 scores on different types of questions. Calculated on the development set.}
\label{type}
\end{table*}

\section{Full results of Performance Analysis over Human-Labeled Question Types}
\label{sec:appA}
Table \ref{tab:reason_results_full} gives our full evaluation on human annotated question types.

Compared with the BiDAF model, one interesting observation is that the generative baseline gets much worse results on ambiguous questions. We conjecture that although these questions are meaningless, they still have many words that overlapped with the contexts. This can give BiDAF potential advantage over the generative baseline.

\section{Performance Analysis over Automatically-Labeled Question Types}
\label{sec:appB}

Besides the analysis on different reasoning types, we also look into the performance over questions with different first tokens in the development set, which provide us an automatic categorization of questions. According to the results in Table~\ref{type}, the three neural baselines all perform the best on ``Who'' and ``Where'' questions, to which the answers are often named entities. Since the tweet contexts are short, there are only a small number of named entities to choose from, which could make the answer pattern easy to learn. On the other hand, the neural models fail to perform well on the ``Why'' questions, and the results of neural baselines are even worse than that of the matching baseline. We find that these questions generally have longer answer phrases than other types of questions, with the average answer length being 3.74 compared to 2.13 for any other types. Also, since all the answers are written by humans instead of just spans from the context, these abstractive answers can make it even harder for current models to handle. We also observe that when people write ``Why'' questions, they tend to copy word spans from the tweet, potentially making the task easier for the matching baseline.

\end{document}